\documentclass{article}

\usepackage[dblblindworkshop,final]{neurips_2026}
\usepackage[utf8]{inputenc}
\usepackage[T1]{fontenc}
\usepackage{hyperref}
\usepackage{url}
\usepackage{booktabs}
\usepackage{amsmath,amssymb,mathtools}
\usepackage{amsthm}
\usepackage{bm}
\newcommand{\mat}[1]{\bm{#1}}

\usepackage{microtype}
\usepackage{xcolor}
\usepackage{graphicx}
\usepackage{caption}
\usepackage{wrapfig}
\usepackage{needspace}
\usepackage{tabularx}
\usepackage{multirow}
\usepackage{colortbl}
\usepackage{pifont}
\newcommand{\xmark}{\ding{55}}%
\usepackage{placeins}
\usepackage{enumitem}

\definecolor{navy}{HTML}{17365D}
\definecolor{linkblue}{HTML}{3F79B7}
\definecolor{oraclebg}{HTML}{F1F5F9}
\definecolor{bestbg}{HTML}{D4EDDA}
\hypersetup{colorlinks=true,citecolor=navy,linkcolor=navy,urlcolor=linkblue}

\title{LoRA on a Budget: Function-Aware Post-Hoc Compression for Adapter Fleets}

\workshoptitle{Neural Network Artifacts as a New Data Modality}

\author{
  Shenghui Li$^{1}$ and Thiemo Voigt$^{1,2}$ \\
  $^{1}$Uppsala University, Uppsala, Sweden \\
  $^{2}$Research Institutes of Sweden, Stockholm, Sweden \\
  \texttt{\{shenghui.li,thiemo.voigt\}@angstrom.uu.se}
}

\begin{document}

\maketitle

\begin{abstract}
Low-Rank Adaptation (LoRA) has become a key mechanism for efficiently
specializing foundation models into many task- and user-specific variants. Serving many LoRA adapters under a shared resource budget turns post-hoc
compression into a two-level allocation problem: rank must be distributed
across layers within each adapter and across adapters within the fleet.
Although trained LoRAs contain substantial spectral redundancy, spectral
reconstruction alone does not determine how much compression an adapter can
functionally tolerate. We formulate fleet compression through a functional rank--distortion frontier
for each adapter, followed by fleet-level minimax allocation under a shared
rank budget. Given exact frontiers, this decomposition is equivalent to joint
optimization over all layer ranks. We introduce \textbf{Functional Rank Allocation (FRA)}, an unlabeled method for constructing practical frontiers.
FRA measures isolated module truncations end to end, uses these measurements as
an additive search surrogate solved by dynamic programming, remeasures every
complete candidate, and takes the lower-risk envelope with a complementary
spectral proposal before exact minimax allocation over the constructed frontiers. Across 78 adapters from three
public fleets, FRA improves worst-case and tail utility over
uniform spectral compression in constrained-budget regimes. Fixed-budget probes
show that isolated costs rank candidates well, but their calibration shifts
across proposal shapes; end-to-end verification and the spectral proposal
therefore limit functional-search error.
\end{abstract}

\section{Introduction}
\label{sec:intro}

Foundation models are increasingly adapted into specialized models for a wide
range of downstream tasks. Updating every parameter of these models, however,
is costly to train, store, and deploy, particularly when many task-specific
variants share the same foundation model. Parameter-efficient fine-tuning
(PEFT) addresses this cost by learning only a small set of task-specific
parameters while leaving the pretrained model frozen. Low-rank adaptation
(LoRA) \citep{hu2022lora} has emerged as an especially effective form of PEFT.
Rather than modifying a weight matrix directly, LoRA learns an additive update
represented as the product of two low-rank factors. This parameterization can
retain the benefits of task-specific fine-tuning with far fewer trainable and
stored parameters, making each specialization a compact adapter that can be
shared, composed with a common base model, and loaded on demand.

A collection of LoRA adapters served from a shared base model forms a
\emph{LoRA fleet}. Modern inference frameworks increasingly support serving
such fleets, with individual adapters selected or loaded on a per-request
basis. Systems such as vLLM \citep{kwon2023vllm},
S-LoRA \citep{sheng2024slora}, Punica \citep{chen2024punica}, and
LoRAX \citep{predibase2023lorax} enable such multi-LoRA serving by batching,
routing, and managing adapters around shared base weights. At this scale,
adapters compete for finite GPU cache capacity, transfer bandwidth, and
rank-dependent inference resources. Compress-then-Serve shows that compressing
an adapter fleet can translate into substantial serving gains
\citep{brueelgabrielsson2024compressthenserve}.

\begin{wrapfigure}{r}{0.48\textwidth}
	\vspace{-1.2\baselineskip}
	\centering
	\includegraphics[width=\linewidth]{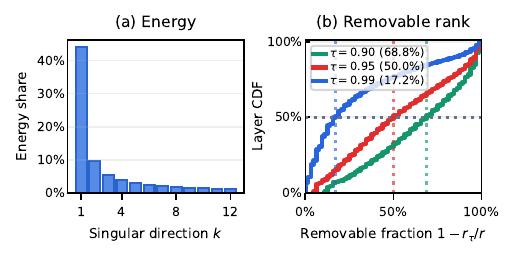}
	\captionsetup{font=footnotesize}
	\caption{\textbf{LoRA fleets contain substantial compressible rank.}
		Among $96{,}370$ matrices, the median layer retains only $31.2\%$ of rank at
		$\tau=.90$.}
	\label{fig:rank-utilization}
	\vspace{-0.6\baselineskip}
\end{wrapfigure}

		There is substantial capacity to reclaim from trained LoRA adapters. To
		characterize rank redundancy in the wild, we analyze $96{,}370$ matrices from
		200 popular LoRA repositories on Hugging Face and find that spectral energy is
		strongly concentrated: at $\tau=.90$, the median layer retains only $31.2\%$
		of its nominal rank (Fig.~\ref{fig:rank-utilization}). This widespread
		redundancy makes post-hoc compression an attractive way to reduce adapter
		memory, transfer volume, and rank-dependent serving overhead, allowing a fixed
		serving capacity to accommodate a larger adapter fleet
		\citep{phlora2025,kumaravelu2026para,
			brueelgabrielsson2024compressthenserve,compressthenmerge2026}. The challenge, however, is no longer simply whether rank can be removed, but
how a limited rank budget should be allocated. Within a single adapter, different layers need not tolerate the same degree of truncation. Accordingly, the available rank should be distributed to preserve the adapter's behavior as well as possible. At fleet scale, the problem becomes coupled, as adapters share the same serving capacity yet may differ substantially in how much compression they can tolerate. Effective post-hoc compression must therefore decide both how rank is divided across layers and how the shared budget is divided across
adapters.

We address this two-level allocation problem with \textbf{Functional Rank
Allocation (FRA)}. For each adapter, the ideal objective is a functional
rank--distortion frontier that captures the minimum behavioral change
achievable at every retained-rank budget. Constructing this frontier exactly is
combinatorial, so FRA uses end-to-end measurements of individual module
truncations to form an additive search surrogate and solves the resulting
rank-allocation problem exactly by dynamic programming. Each proposed
allocation is then instantiated and remeasured end to end, so the surrogate is
used only to generate candidates rather than to assign their final risk.
Because the additive surrogate can be calibrated differently across allocation
shapes, FRA also evaluates a complementary spectral candidate and retains the lower-risk
allocation at each budget. The resulting measured frontiers are finally
coupled through exact minimax allocation over these constructed frontiers under
the shared rank budget. We summarize our contribution as follows:
\begin{itemize}[leftmargin=1.5em,labelsep=0.5em]
  \item We formulate post-hoc LoRA fleet compression as a two-level resource
  allocation problem: each adapter first defines a functional
  rank--distortion frontier over its layer-wise allocations, after which the
  shared fleet budget is allocated across these frontiers. Separability makes
  this decomposition equivalent to joint layer-rank optimization when the
  frontiers are exact; the practical candidate frontiers are approximate.

  \item We introduce \textbf{Functional Rank Allocation (FRA)}, an unlabeled
  method for constructing practical functional frontiers. FRA uses isolated
  end-to-end truncation costs to search for layer-rank allocations via dynamic
  programming, remeasures every complete candidate before it enters the
  frontier, and adds a spectral candidate to guard against cross-proposal
  surrogate miscalibration.

  \item We evaluate FRA across 78 adapters from three public fleets and show
  that function-aware allocation substantially improves tail utility under
  constrained budgets. We show that isolated costs preserve within-budget
  ordering, while proposal-shape-dependent calibration error explains why the
  complementary spectral candidate remains valuable.
\end{itemize}

\section{Background}
\label{sec:background}

\paragraph{Post-Hoc LoRA Compression}
For a pretrained weight matrix
$\mat W_{0,\ell}\in\mathbb{R}^{d_{\mathrm{out}}\times d_{\mathrm{in}}}$,
LoRA keeps $\mat W_{0,\ell}$ frozen and learns an additive low-rank update
$
  \Delta\mat W_{i,\ell}
  =
  \frac{\alpha_{i,\ell}}{r_{i,\ell}}
  \mat B_{i,\ell}\mat A_{i,\ell},
  \label{eq:lora-update}
$
where
$\mat B_{i,\ell}\in\mathbb{R}^{d_{\mathrm{out}}\times r_{i,\ell}}$,
$\mat A_{i,\ell}\in\mathbb{R}^{r_{i,\ell}\times d_{\mathrm{in}}}$,
and $r_{i,\ell}$ is the nominal rank \citep{hu2022lora}.
An adapter consists of such updates across a collection of modules while
sharing the same frozen base model. An already-trained LoRA module can be further compressed without retraining by
reducing the rank of its effective update $\Delta\mat W_{i,\ell}$. Let its
compact singular value decomposition be
$
  \Delta\mat W_{i,\ell}
  =
  \mat U_{i,\ell}
  \mat\Sigma_{i,\ell}
  \mat V_{i,\ell}^{\top},
  \label{eq:lora-svd}
$
with singular values
$\sigma_{i,\ell,1}\ge\cdots\ge\sigma_{i,\ell,r_{i,\ell}}$.
Retaining the leading $k$ singular directions gives
$
  \Delta\mat W_{i,\ell,k}
  =
  \mat U_{i,\ell,:k}
  \mat\Sigma_{i,\ell,:k}
  \mat V_{i,\ell,:k}^{\top},
  \label{eq:lora-truncation}
$
the optimal rank-$k$ approximation under the Frobenius norm. The truncated
update can be refactored into rank-$k$ LoRA factors and served in the same form
as the original module. Such SVD recompression is used in post-hoc adapter
compression \citep{brueelgabrielsson2024compressthenserve,kumaravelu2026para}.
It can also be computed directly from low-rank factors or compact coordinate
matrices without materializing the full update, substantially reducing the
recompression cost \citep{kumaravelu2026para,li2026fraq}. A common spectral compression rule selects the smallest retained rank whose
leading singular directions preserve a prescribed fraction $\tau$ of the
spectral energy,
$
  k_{i,\ell}(\tau)
  =
  \min\left\{
  k\in\{1,\ldots,r_{i,\ell}\}:
  \frac{\sum_{j=1}^{k}\sigma_{i,\ell,j}^{2}}
       {\sum_{j=1}^{r_{i,\ell}}\sigma_{i,\ell,j}^{2}}
  \ge \tau
  \right\}.
  \label{eq:spectral-rank}
$
We additionally allow $k=0$, which removes the corresponding LoRA branch
entirely.

\paragraph{LoRA Fleets and Serving Budgets}
For a fleet of $N$ adapters sharing a base model, let $k_{i,\ell}$ denote the
retained rank of module $\ell$ in adapter $i$ after post-hoc compression and
$
  K_i=\sum_{\ell}k_{i,\ell}
  \label{eq:adapter-rank-budget}
$
its aggregate retained rank. Within each fleet, architecture and module shapes
are fixed across adapters. We report a hardware-agnostic budget in retained
rank-direction slots and impose
$
  \sum_{i=1}^{N}K_i\le B,
  \label{eq:fleet-rank-budget}
$
where $B$ is a direction budget rather than an exact byte count. Modules of
different width have different per-direction storage cost; a deployment can
use weighted costs $c_\ell k_{i,\ell}$, with
$c_\ell=d_{\mathrm{in},\ell}+d_{\mathrm{out},\ell}$ parameters per direction,
within the same optimization formulation.

\section{Method}
\label{sec:method}

\subsection{Problem Formulation}

Given the fleet setting above, post-hoc LoRA compression requires deciding how
capacity should be distributed across layers within each adapter and across
adapters sharing the same serving capacity. We formulate both allocation
problems under a common functional risk that quantifies how compression changes
model behavior.

\textbf{Single-Adapter Compression.}
Consider adapter $i$ with $L_i$ LoRA modules and layer-wise rank allocation
$\mathbf{k}_i=(k_{i,1},\ldots,k_{i,L_i})$, where
$k_{i,\ell}\in\{0,\ldots,r_{i,\ell}\}$ and $k_{i,\ell}=0$ removes that LoRA
branch from execution. Let $f_i$ denote the uncompressed
adapter and $f_{i,\mathbf{k}_i}$ its compressed counterpart. We define
$R_i(\mathbf{k}_i)=
\mathbb{E}_{x\sim\mathcal{C}_i}
[\phi(f_i(x),f_{i,\mathbf{k}_i}(x))]$,
where $\mathcal{C}_i$ is a small unlabeled calibration set and $\phi$ measures
behavioral change. By default, we use output-distribution Jensen--Shannon (JS) divergence.
This is an unlabeled behavioral proxy, not a guarantee on downstream utility,
which we evaluate separately on held-out labeled data.
For an adapter-level rank budget $K$, the ideal compression defines the
functional rank--distortion frontier
\begin{equation}
	R_i^\star(K)
	=
	\min_{\|\mathbf{k}_i\|_1\le K} R_i(\mathbf{k}_i),
	\qquad
	\mathbf{k}_i^\star(K)
	\in
	\arg\min_{\|\mathbf{k}_i\|_1\le K} R_i(\mathbf{k}_i).
	\label{eq:single-frontier}
\end{equation}

\textbf{Multi-Adapter Compression.}
For a fleet of $N$ adapters sharing a total rank budget $B$, we allocate an
adapter-level budget $K_i$ to each adapter by solving
\begin{equation}
	\min_{K_1,\ldots,K_N}
	\max_{i=1,\ldots,N} R_i^\star(K_i)
	\quad
	\mathrm{s.t.}\quad
	\sum_{i=1}^{N}K_i\le B.
	\label{eq:fleet-objective}
\end{equation}
Note that we use the minimax objective to prevent a small subset of adapters from
absorbing disproportionate compression damage. Other objectives, such as mean
or workload-weighted risk, can be optimized over the same frontiers. Since
each adapter's risk depends only on its own ranks, using exact frontiers
$R_i^\star(K)$ makes Eq.~\eqref{eq:fleet-objective} equivalent to joint
optimization over all layer ranks under the global budget.

\subsection{Functional Rank Allocation}
\label{sec:functional-rank-allocation}

Constructing the exact frontier $R_i^\star(K)$, however, requires searching
over a combinatorial space of layer-wise rank allocations. FRA approximates
this search using functional measurements of individual truncation actions,
while remeasuring each complete candidate end to end. We probe each compression
action in isolation. For
module $\ell$, let $  C_{i,\ell}(k)
  =R_i\!\left(\mathbf{k}_i^{(\ell\rightarrow k)}\right)
  \label{eq:isolated-cost} $
denote the functional risk for the rank vector
$\mathbf{k}_i^{(\ell\rightarrow k)}$, which truncates only module $\ell$ to rank $k$
with all other modules left uncompressed. Each $C_{i,\ell}(k)$ is an
end-to-end functional measurement on unlabeled calibration prompts, rather
than a spectral or activation-based proxy.

We use the sum of isolated costs only as an additive search surrogate,
$\widetilde R_i(\mathbf{k}_i)=\sum_{\ell=1}^{L_i}
C_{i,\ell}(k_{i,\ell})$. For budget $K$, we solve
\begin{equation}
  \widehat{\mathbf{k}}_i^{\mathrm F}(K)
  =\arg\min_{\|\mathbf{k}_i\|_1\le K}
  \sum_{\ell=1}^{L_i}C_{i,\ell}(k_{i,\ell}).
  \label{eq:funcdp}
\end{equation}
Eq.~\eqref{eq:funcdp} is a multiple-choice knapsack problem, where each
module forms one choice group and the retained rank is its resource cost.
We solve it exactly for the surrogate objective by dynamic programming in
$O(L_i K r_{\max})$ time, where $r_{\max}$ is the largest candidate rank.

The additive surrogate is used only to search for a rank allocation. For
every proposed $\widehat{\mathbf{k}}_i^{\mathrm F}(K)$, we instantiate the
complete compressed adapter and remeasure its end-to-end functional risk,
$
  \widehat R_i^{\mathrm F}(K)
  =R_i\!\left(\widehat{\mathbf{k}}_i^{\mathrm F}(K)\right).
  \label{eq:functional-verification}
$
Thus, approximation errors affect which allocation is proposed, but not the
risk value exposed to the fleet-level optimizer. %
The surrogate can incur search error when isolated effects interact. We retain
a spectral proposal independent of the additive functional surrogate,
$\widehat{\mathbf{k}}_i^{\mathrm S}(K)\in\arg\min_{\|\mathbf{k}\|_1\le K}
\sum_\ell\|\Delta W_{i,\ell}-\Delta W_{i,\ell,k_\ell}\|_F^2$, equivalently
keeping the largest prefix-constrained singular energies. We evaluate it with the
same end-to-end risk,
$\widehat R_i^{\mathrm S}(K)=
R_i(\widehat{\mathbf{k}}_i^{\mathrm S}(K))$, and retain the lower measured risk
at each budget, i.e.,
$  \widehat R_i(K)
  =\min\!\left\{
    \widehat R_i^{\mathrm F}(K),
    \widehat R_i^{\mathrm S}(K)
  \right\}.
  \label{eq:robust-frontier}
$
We denote the corresponding retained rank vector by
$\widehat{\mathbf{k}}_i(K)$. This measured curve approximates the ideal
frontier $R_i^\star(K)$; it need not attain the optimum over all rank vectors.
Let $\mathcal{G}_i$ denote the finite grid of adapter budgets at which we
construct these candidates.

\subsection{Fleet-Level Minimax Allocation}
\label{sec:fleet-minimax-allocation}

After Sec.~\ref{sec:functional-rank-allocation}, each adapter $i$ is represented
by a measured functional frontier $\widehat R_i(K)$ together with the
corresponding layer-rank allocation. Because the global constraint is an upper
budget, we take the monotone closure
$\overline R_i(K)=\min_{K'\le K}\widehat R_i(K')$: if a smaller allocation
achieves lower risk, there is no reason to spend additional rank. The
corresponding rank vector is retained for deployment. Given these frontiers, we solve
\begin{equation}
  \min_{K_1,\ldots,K_N}\max_i\overline R_i(K_i)
  \quad\mathrm{s.t.}\quad \sum_{i=1}^N K_i\le B.
  \label{eq:fleet-minimax}
\end{equation}
For any candidate risk level $\tau$, adapter $i$ requires at least
$K_i(\tau)=\min\{K:\overline R_i(K)\le\tau\}$ rank to remain below $\tau$.
Thus, $\tau$ is feasible if and only if $\sum_i K_i(\tau)\le B$. We enumerate
the finite set of measured risk levels and select the smallest feasible
$\tau$, yielding the exact minimax solution over the constructed frontiers. Among allocations attaining the same minimax value, we minimize total risk and
then maximize budget utilization.
For a sorted grid of $G_i=|\mathcal{G}_i|$ budgets, monotone closure is a
prefix minimum and takes $O(G_i)$ time. Writing $H=\sum_i G_i$, a simple
implementation sorts the measured risk levels in $O(H\log H)$ time and
checks all levels in $O(H^2)$ time by scanning the frontiers; these bounds
concern the minimax value, before tie-breaking. Frontier measurement requires
$\sum_\ell r_{i,\ell}+2G_i$ calibration-set evaluations per adapter for FRA,
excluding a reusable uncompressed reference.

\begin{table}[t]
  \captionsetup{font=footnotesize}
  \caption{\textbf{Fleet scale and what each budget costs.} $N$ is the number of
  adapters in each pool and $r$ is the nominal rank. Grouped-query attention makes a $q$ direction cost $8{,}192$
  parameters against $5{,}120$ for $k/v$, so on the Mistral fleets the parameter
  share departs from the rank share.}
  \label{tab:budgets}
  \small
  \setlength{\tabcolsep}{4.2pt}
  \renewcommand{\arraystretch}{1.15}
  \aboverulesep=0ex
  \belowrulesep=0ex
  \centering
  \resizebox{\linewidth}{!}{%
  \begin{tabular}{c | ccc | c | c | c | c | c}
    \toprule
    \multicolumn{4}{c|}{\textbf{Pool}} & \multirow{2}{*}{\textbf{Metric}} & \multirow{2}{*}{\textbf{Full rank}} & \textbf{Moderate} & \textbf{Binding} & \textbf{Severe} \\
    \cline{1-4}
    Fleet & $r$ & $N$ & Modules & & & ($\tau_{\mathrm{uni}}{=}.90$) & ($\tau_{\mathrm{uni}}{=}.70$) & ($\tau_{\mathrm{uni}}{=}.50$) \\
    \midrule
    \multirow{2}{*}{\textbf{LoRA Land}} & \multirow{2}{*}{$8$} & \multirow{2}{*}{$12$} & \multirow{2}{*}{$64\times$(q,v)}
      & Directions & $6{,}144$ & $2{,}288$ ($37.2\%$) & $1{,}320$ ($21.5\%$) & $936$ ($15.2\%$) \\
      & & & & Params (M) & $40.89$ & $15.28$ ($37.4\%$) & $8.70$ ($21.3\%$) & $6.09$ ($14.9\%$) \\
    \midrule
    \multirow{2}{*}{\textbf{Lots-of-LoRAs}} & \multirow{2}{*}{$16$} & \multirow{2}{*}{$25$} & \multirow{2}{*}{$96\times$(q,k,v)}
      & Directions & $38{,}400$ & $19{,}200$ ($50.0\%$) & $8{,}800$ ($22.9\%$) & $4{,}912$ ($12.8\%$) \\
      & & & & Params (M) & $235.93$ & $119.59$ ($50.7\%$) & $55.57$ ($23.6\%$) & $30.50$ ($12.9\%$) \\
    \midrule
    \multirow{2}{*}{\textbf{LoRARetriever}} & \multirow{2}{*}{$8$} & \multirow{2}{*}{$41$} & \multirow{2}{*}{$64\times$(q,v)}
      & Directions & $20{,}992$ & $5{,}880$ ($28.0\%$) & $3{,}184$ ($15.2\%$) & $2{,}680$ ($12.8\%$) \\
      & & & & Params (M) & $171.97$ & $48.17$ ($28.0\%$) & $26.08$ ($15.2\%$) & $21.95$ ($12.8\%$) \\
    \bottomrule
  \end{tabular}%
  }
\end{table}

\begin{table}[t]
  \centering
  \captionsetup{font=footnotesize}
  \caption{\textbf{Nested ablation at matched fleet rank-direction budgets.}
  Uniform $\rightarrow$ Intra-Spec $\rightarrow$ E2E-Spec $\rightarrow$
  FuncDP $\rightarrow$ FRA progressively adds within-adapter reallocation,
  cross-adapter reallocation, functional inner search, and the spectral
  safeguard. The \emph{Candidate Oracle} gives, per statistic, the best value any
  allocation in the union of the five families reaches at that budget, so a
  block's entries need not share one allocation. FRA is bold; green marks the
  best deployable entry.}
  \label{tab:main}
  \small
  \renewcommand{\arraystretch}{1.02}
  \setlength{\tabcolsep}{2.9pt}
  \aboverulesep=0ex
  \belowrulesep=0ex
  \resizebox{\linewidth}{!}{%
\begin{tabular}{l | l | c | ccc | ccc | ccc}
    \toprule
    \multirow{2}{*}{\textbf{Pool}} & \multirow{2}{*}{\textbf{Method}} & \multirow{2}{*}{\textbf{Probes}} & \multicolumn{3}{c|}{\textbf{Moderate} ($\tau_{\mathrm{uni}}{=}.90$)} & \multicolumn{3}{c|}{\textbf{Binding} ($\tau_{\mathrm{uni}}{=}.70$)} & \multicolumn{3}{c}{\textbf{Severe} ($\tau_{\mathrm{uni}}{=}.50$)} \\
    \cline{4-12}
    & & & Mean & $p_{10}$ & Worst & Mean & $p_{10}$ & Worst & Mean & $p_{10}$ & Worst \\
    \midrule
    \multirow{6}{*}{\textbf{LoRA Land}}
    & \cellcolor{oraclebg}\textit{Cand. Oracle} & \cellcolor{oraclebg}\textit{labeled} & \cellcolor{oraclebg}\textit{1.009} & \cellcolor{oraclebg}\textit{0.990} & \cellcolor{oraclebg}\textit{0.966} & \cellcolor{oraclebg}\textit{0.983} & \cellcolor{oraclebg}\textit{0.949} & \cellcolor{oraclebg}\textit{0.879} & \cellcolor{oraclebg}\textit{0.921} & \cellcolor{oraclebg}\textit{0.878} & \cellcolor{oraclebg}\textit{0.672} \\
    \cline{2-12}
    & Uniform & \xmark & \phantom{-}$0.913$ & \phantom{-}$0.873$ & \phantom{-}$0.081$ & \phantom{-}$0.725$ & \phantom{-}$0.010$ & $-0.293$ & \phantom{-}$0.748$ & \phantom{-}$0.047$ & $-0.015$ \\
    & Intra-Spec & \xmark & \phantom{-}$0.895$ & \phantom{-}$0.898$ & \phantom{-}$0.036$ & \phantom{-}$0.764$ & \phantom{-}$0.185$ & \phantom{-}$0.000$ & \phantom{-}$0.554$ & $-0.011$ & $-0.414$ \\
    & E2E-Spec & curve & \phantom{-}$0.974$ & \phantom{-}$0.920$ & \phantom{-}$0.897$ & \phantom{-}$0.829$ & \phantom{-}$0.640$ & \phantom{-}$0.203$ & \phantom{-}$0.530$ & $-0.030$ & $-1.309$ \\
    & FuncDP & curve\,+\,layer & \cellcolor{bestbg}\phantom{-}$0.985$ & \cellcolor{bestbg}\phantom{-}$0.949$ & \cellcolor{bestbg}\phantom{-}$0.931$ & \cellcolor{bestbg}\phantom{-}$0.963$ & \cellcolor{bestbg}\phantom{-}$0.915$ & \cellcolor{bestbg}\phantom{-}$0.879$ & \cellcolor{bestbg}\phantom{-}$0.907$ & \cellcolor{bestbg}\phantom{-}$0.692$ & \cellcolor{bestbg}\phantom{-}$0.620$ \\
    & \textbf{FRA} & $2\times$curve\,+\,layer & \cellcolor{bestbg}\phantom{-}$0.985$ & \cellcolor{bestbg}\phantom{-}$0.949$ & \cellcolor{bestbg}\phantom{-}$0.931$ & \cellcolor{bestbg}\phantom{-}$0.963$ & \cellcolor{bestbg}\phantom{-}$0.915$ & \cellcolor{bestbg}\phantom{-}$0.879$ & \cellcolor{bestbg}\phantom{-}$0.907$ & \cellcolor{bestbg}\phantom{-}$0.692$ & \cellcolor{bestbg}\phantom{-}$0.620$ \\
    \midrule[\heavyrulewidth]
    \multirow{6}{*}{\textbf{Lots-of-LoRAs}}
    & \cellcolor{oraclebg}\textit{Cand. Oracle} & \cellcolor{oraclebg}\textit{labeled} & \cellcolor{oraclebg}\textit{1.033} & \cellcolor{oraclebg}\textit{1.000} & \cellcolor{oraclebg}\textit{1.000} & \cellcolor{oraclebg}\textit{1.017} & \cellcolor{oraclebg}\textit{0.974} & \cellcolor{oraclebg}\textit{0.955} & \cellcolor{oraclebg}\textit{0.998} & \cellcolor{oraclebg}\textit{0.923} & \cellcolor{oraclebg}\textit{0.877} \\
    \cline{2-12}
    & Uniform & \xmark & \phantom{-}$0.989$ & \phantom{-}$0.960$ & \phantom{-}$0.877$ & \phantom{-}$0.968$ & \phantom{-}$0.885$ & \phantom{-}$0.542$ & \phantom{-}$0.921$ & \phantom{-}$0.797$ & \phantom{-}$0.351$ \\
    & Intra-Spec & \xmark & \phantom{-}$0.998$ & \phantom{-}$0.954$ & \phantom{-}$0.927$ & \phantom{-}$0.961$ & \phantom{-}$0.882$ & \phantom{-}$0.515$ & \phantom{-}$0.892$ & \phantom{-}$0.469$ & \phantom{-}$0.321$ \\
    & E2E-Spec & curve & \phantom{-}$0.998$ & \phantom{-}$0.977$ & \phantom{-}$0.895$ & \phantom{-}$0.974$ & \phantom{-}$0.915$ & \phantom{-}$0.599$ & \phantom{-}$0.961$ & \phantom{-}$0.879$ & \cellcolor{bestbg}\phantom{-}$0.740$ \\
    & FuncDP & curve\,+\,layer & \cellcolor{bestbg}\phantom{-}$1.004$ & \cellcolor{bestbg}\phantom{-}$0.988$ & \phantom{-}$0.910$ & \phantom{-}$0.989$ & \phantom{-}$0.934$ & \phantom{-}$0.800$ & \cellcolor{bestbg}\phantom{-}$0.973$ & \cellcolor{bestbg}\phantom{-}$0.912$ & \phantom{-}$0.599$ \\
    & \textbf{FRA} & $2\times$curve\,+\,layer & \phantom{-}$1.000$ & \phantom{-}$0.982$ & \cellcolor{bestbg}\phantom{-}$0.935$ & \cellcolor{bestbg}\phantom{-}$1.000$ & \cellcolor{bestbg}\phantom{-}$0.949$ & \cellcolor{bestbg}\phantom{-}$0.910$ & \phantom{-}$0.971$ & \cellcolor{bestbg}\phantom{-}$0.912$ & \phantom{-}$0.599$ \\
    \midrule[\heavyrulewidth]
    \multirow{6}{*}{\textbf{LoRARetriever}}
    & \cellcolor{oraclebg}\textit{Cand. Oracle} & \cellcolor{oraclebg}\textit{labeled} & \cellcolor{oraclebg}\textit{0.986} & \cellcolor{oraclebg}\textit{0.897} & \cellcolor{oraclebg}\textit{0.750} & \cellcolor{oraclebg}\textit{0.933} & \cellcolor{oraclebg}\textit{0.784} & \cellcolor{oraclebg}\textit{0.500} & \cellcolor{oraclebg}\textit{0.874} & \cellcolor{oraclebg}\textit{0.667} & \cellcolor{oraclebg}\textit{0.250} \\
    \cline{2-12}
    & Uniform & \xmark & \phantom{-}$0.905$ & \phantom{-}$0.714$ & \phantom{-}$0.500$ & \phantom{-}$0.819$ & \phantom{-}$0.500$ & \phantom{-}$0.250$ & \phantom{-}$0.791$ & \phantom{-}$0.500$ & \cellcolor{bestbg}\phantom{-}$0.250$ \\
    & Intra-Spec & \xmark & \cellcolor{bestbg}\phantom{-}$0.948$ & \cellcolor{bestbg}\phantom{-}$0.818$ & \cellcolor{bestbg}\phantom{-}$0.600$ & \phantom{-}$0.840$ & \phantom{-}$0.571$ & \phantom{-}$0.143$ & \phantom{-}$0.776$ & \phantom{-}$0.364$ & \phantom{-}$0.143$ \\
    & E2E-Spec & curve & \phantom{-}$0.930$ & \cellcolor{bestbg}\phantom{-}$0.818$ & \phantom{-}$0.500$ & \phantom{-}$0.857$ & \phantom{-}$0.680$ & \cellcolor{bestbg}\phantom{-}$0.500$ & \phantom{-}$0.780$ & \phantom{-}$0.500$ & \cellcolor{bestbg}\phantom{-}$0.250$ \\
    & FuncDP & curve\,+\,layer & \phantom{-}$0.906$ & \phantom{-}$0.667$ & \phantom{-}$0.500$ & \phantom{-}$0.833$ & \phantom{-}$0.600$ & \phantom{-}$0.250$ & \phantom{-}$0.822$ & \cellcolor{bestbg}\phantom{-}$0.600$ & \cellcolor{bestbg}\phantom{-}$0.250$ \\
    & \textbf{FRA} & $2\times$curve\,+\,layer & \phantom{-}$0.933$ & \phantom{-}$0.800$ & \phantom{-}$0.500$ & \cellcolor{bestbg}\phantom{-}$0.894$ & \cellcolor{bestbg}\phantom{-}$0.727$ & \cellcolor{bestbg}\phantom{-}$0.500$ & \cellcolor{bestbg}\phantom{-}$0.829$ & \cellcolor{bestbg}\phantom{-}$0.600$ & \cellcolor{bestbg}\phantom{-}$0.250$ \\
    \bottomrule
\end{tabular}
}
\end{table}

\section{Experiments}
\label{sec:experiments}

We evaluate FRA and competing allocation rules across three diverse
multi-adapter fleets. \textbf{LoRA Land} ($N=12$, Mistral-7B
\citep{predibase2024loraland}) contains 12 reasoning, classification, and
generation tasks, including GSM8K, WikiSQL, and DBpedia, scored with standard
task-specific metrics such as accuracy, BLEU, and Rouge; \textbf{Lots-of-LoRAs}
($N=25$, Mistral-7B-Instruct-v0.2 \citep{lotsofloras2024collection}) contains
25 specialized domain adapters---of the 33 controlled-pool members, three have
no public evaluation dataset and five leave a held-out split too small to score,
so 25 is the full reachable pool rather than a sample; and
\textbf{LoRARetriever} ($N=41$, Llama-2-7B
\citep{zhao2024loraretriever}) contains 41 classification and GLUE/SuperGLUE
adapters.

\Needspace{14\baselineskip}
\begin{wraptable}{r}{0.48\textwidth}
\vspace{-0.5\baselineskip}
	\centering
	\captionsetup{font=footnotesize}
	\caption{\textbf{Calibration cost per adapter.} Each probe evaluates 16
        unlabeled prompts. Counts exclude a reusable uncompressed reference;
        all probes are offline.}
    \label{tab:cost}
	\footnotesize
	\setlength{\tabcolsep}{4pt}\renewcommand{\arraystretch}{0.95}
	\resizebox{\linewidth}{!}{%
\begin{tabular}{@{}lll@{}}
		\toprule
		Rule & Probes & What they measure \\
		\midrule
		Uniform & $0$ & --- \\
		Intra-Spec & $0$ & --- \\
		E2E-Spec & $|\mathcal{G}|$ & each spectral proposal \\
		FuncDP & $Lr+|\mathcal{G}|$ & ranks $0{:}r{-}1$, then proposals \\
		FRA & $Lr+2|\mathcal{G}|$ & both proposal families \\
		\bottomrule
	\end{tabular}%
}
	\end{wraptable}

\paragraph{Calibration and its cost.} No rule except the Candidate Oracle sees a
downstream label.
Calibration prompts are disjoint from the scored examples in every pool
(LoRA Land: examples 200--215 calibrate, 0--199 score; Lots-of-LoRAs: train rows
0--15 calibrate, the held-out split scores; LoRARetriever: eval-set rows 0--15
calibrate, 16--49 score). What differs is how many calibration-set evaluations each
rule needs per adapter (Table~\ref{tab:cost}).
Here $L$ is the module count, $r$ the nominal rank, and
$|\mathcal{G}|$ the number of budgets on the measured grid: $65$
for LoRA Land, $97$ for Lots-of-LoRAs, and $65$ for LoRARetriever. The first two rungs
need no calibration forward passes; their allocations are decided from the spectra. FRA's $Lr$ single-layer probes ($512$ on LoRA Land, with
$L{=}64$ modules and $r{=}8$) dominate its cost, but are paid once per adapter at
registration and reused at every budget; on one GH200 that is about $8$ minutes
per adapter, and nothing is added to the serving path.

\paragraph{Metric and budgets.}
For a task score $M$ oriented so that higher is better, we report retention
$\mathrm{Ret}=\frac{M(\mathrm{compressed})-M(\mathrm{base})}
{M(\mathrm{uncompressed})-M(\mathrm{base})}$.
Higher retention is better: $1$ preserves the uncompressed adapter's gain,
values above $1$ exceed it, and values at or below $0$ do not improve on the
base model. We report the fleet mean, 10th percentile ($p_{10}$), and minimum.
Fleet direction budgets equal the cost of uniform truncation at thresholds
$\tau_{\mathrm{uni}}=.90$, $.70$, and $.50$, defining moderate, binding, and
severe regimes. All methods in Table~\ref{tab:main} share the same direction
budget, not necessarily the same parameter count. Table~\ref{tab:budgets}
reports the parameter costs of the uniform reference allocations; alternative
layer-rank allocations can have different parameter costs under GQA.

\paragraph{Measured Functional Frontiers Matter.}
\label{sec:measured-frontiers}

Spectral information can propose a layer-rank allocation, but the spectrum
alone does not provide the end-to-end risk value required by the fleet
allocator. This distinction matters at fleet scale, where the outer allocator
compares adapters through their frontier values rather than through their
spectra. We therefore separate proposal generation from frontier evaluation:
spectral and functional search mechanisms propose candidate allocations, but
every complete candidate is remeasured end to end before entering the fleet
optimizer.

\paragraph{Ablating the FRA Design.}
Table~\ref{tab:main} relaxes the restrictions of uniform spectral compression
one rung at a time at a fixed fleet direction budget. Uniform applies a shared
spectral-energy threshold. Intra-Spec redistributes rank across modules by
spectral importance while retaining the adapter-level budgets from Uniform.
E2E-Spec uses the spectral proposals $\widehat{\mathbf{k}}_i^{\mathrm S}(K)$
and their measured risks, freeing the adapter-level budgets for the outer
minimax optimizer. FuncDP instead uses the functional proposals
$\widehat{\mathbf{k}}_i^{\mathrm F}(K)$ and their measured risks. FRA combines
both through $\widehat R_i(K)=\min\{\widehat R_i^{\mathrm F}(K),
\widehat R_i^{\mathrm S}(K)\}$ before the same outer optimization. From Intra-Spec onward
modules may receive rank zero, and the Candidate Oracle is a reference over the
same candidate families rather than a rung. Within-adapter spectral
reallocation alone is inconsistent, whereas releasing adapter-level budgets
helps more when compression sensitivity differs across tenants. At the binding
budget on LoRA Land, worst-case retention rises from $-0.293$ under Uniform to
$0.203$ with E2E-Spec, and functional inner allocation further raises it to
$0.879$; on Lots-of-LoRAs, FuncDP improves the corresponding worst case from
$0.599$ to $0.800$. The spectral envelope then
protects against failures of the functional surrogate: FRA preserves the
$0.879$ worst case on LoRA Land, raises Lots-of-LoRAs from $0.800$ to $0.910$,
and restores LoRARetriever from $0.250$ to $0.500$.
FRA weakly dominates either proposal family in measured functional risk at each
constructed adapter budget, while Table~\ref{tab:main} evaluates downstream
utility on held-out data. On LoRA Land, adding the spectral envelope leaves
the reported retention unchanged at the displayed precision.

\Needspace{17\baselineskip}
\begin{wrapfigure}{r}{0.48\textwidth}
\vspace{-0.5\baselineskip}

	\centering
	\includegraphics[width=\linewidth]{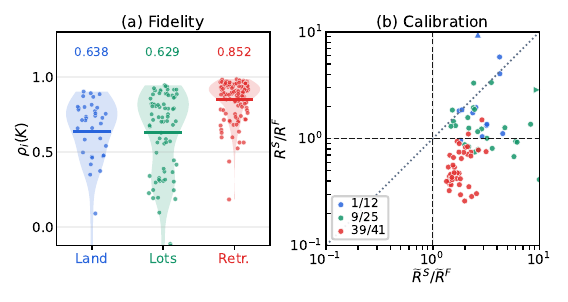}
	\captionsetup{font=footnotesize}
	\caption{\textbf{Ranking fidelity does not explain the cross-proposal
		reversal.} \textbf{(a)} fixed-budget Spearman fidelity, fleet means
		above. \textbf{(b)} per-adapter median surrogate versus measured risk
		ratio on a shared $10^{-1}$--$10^{1}$ scale, colors as in (a); the legend
		counts spectral-favoring adapters and triangles mark the two adapters
		beyond the axes.}
	\label{fig:frontier}

\end{wrapfigure}

\paragraph{When Does Inner Allocation Help?}
The dynamic programming algorithm solves the surrogate optimization problem in Eq.~\eqref{eq:funcdp} exactly.
To assess how well the surrogate ranks candidate allocations, we measure within-budget Spearman rank correlation between surrogate and measured risks.
Here $P$ denotes a parameter budget, distinct from the direction budget $K$.
Let $\mathcal{A}_i^{\mathrm{eval}}(P)\subseteq
\{\mathbf{k}:\sum_\ell c_\ell k_\ell=P\}$ be the finite set of allocations
actually evaluated in the fixed-budget probes, rather than the entire
combinatorial set of feasible allocations. For this set, let $\widetilde{\mathbf r}$ and $\mathbf r$ collect the
surrogate and measured risks in the same order. Ranking fidelity is their
Spearman correlation, $\rho_i(P)=\operatorname{corr}_{\mathrm S}
(\widetilde{\mathbf r},\mathbf r)$.
The means over the evaluated adapter--budget pairs are $0.638$, $0.629$, and
$0.852$ for the three fleets, respectively (Fig.~\ref{fig:frontier}a).
These describe ordering within the evaluated sets, not exhaustive fidelity
across all possible allocations. LoRARetriever has the highest mean fidelity.
The failure instead appears across proposal shapes. Matching functional and
spectral proposals to within $2\%$ parameter cost (excluding identical
allocations), measurement favors functional search in $73.7\%$, $58.0\%$, and
$28.3\%$ of 585, 2,038, and 1,770 comparisons. More independently, the median
measured ratio favors spectral for $1/12$, $9/25$, and $39/41$ adapters.
Median surrogate/measurement spectral-to-functional risk ratios are
$2.74/1.73$, $3.02/1.15$, and $1.76/0.56$: miscalibration grows from $1.6\times$
to $3.1\times$ and reverses the measured sign on LoRARetriever
(Fig.~\ref{fig:frontier}b). This
proposal-shape-dependent surrogate miscalibration is consistent with
shape-dependent cross-layer cancellation, but does not identify it over
higher-order interactions, allocation-severity effects, or selection-induced
optimism. End-to-end remeasurement and the spectral envelope limit this error.

\FloatBarrier
\section{Conclusion}
\label{sec:conclusion}

We presented FRA for function-aware post-hoc rank allocation within and
across LoRA adapters. Isolated truncation probes generate candidates through
dynamic programming; end-to-end remeasurement and a spectral envelope form
approximate functional frontiers. Fleet allocation is exact over those
constructed frontiers, not over all possible layer-rank allocations.
Experiments across three public fleets show improved tail utility in
constrained-budget regimes and identify cross-proposal surrogate
miscalibration as a reason to retain complementary spectral candidates.

\begin{ack}
Computational resources were provided on the Arrhenius system by the National
Academic Infrastructure for Supercomputing in Sweden (NAISS), funded by the
Swedish Research Council.
\end{ack}

\end{document}